\documentclass[letterpaper]{article} 
\usepackage{aaai25}  
\usepackage{times}  
\usepackage{helvet}  
\usepackage{courier}  
\usepackage[hyphens]{url}  
\usepackage{graphicx} 
\usepackage{natbib}  
\usepackage{caption} 
\usepackage{algorithm}
\usepackage{algorithmic}

\usepackage{newfloat}
\usepackage{listings}
\DeclareCaptionStyle{ruled}{labelfont=normalfont,labelsep=colon,strut=off} 
\floatstyle{ruled}
\newfloat{listing}{tb}{lst}{}
\floatname{listing}{Listing}
\usepackage{amsmath}
\usepackage{amsfonts}
\usepackage{cleveref}
\usepackage{booktabs}
\newcommand{\tb}[1]{{\textbf{#1}}}

\title{Decoupling Appearance Variations with 3D Consistent Features\\ in Gaussian Splatting}
\author {
	Jiaqi Lin\textsuperscript{\rm 1},\;
	Zhihao Li\textsuperscript{\rm 2},\;
	Binxiao Huang\textsuperscript{\rm 3},\;
	Xiao Tang\textsuperscript{\rm 2},\;
	Jianzhuang Liu\textsuperscript{\rm 4},\\
	Shiyong Liu\textsuperscript{\rm 2},\;
	Xiaofei Wu\textsuperscript{\rm 2},\;
	Fenglong Song\textsuperscript{\rm 2},\;
	Wenming Yang\textsuperscript{\rm 1}\thanks{Corresponding author.}
}
\affiliations {
	\textsuperscript{\rm 1}Tsinghua University\\
	\textsuperscript{\rm 2}Huawei Noah's Ark Lab\\
	\textsuperscript{\rm 3}The University of Hong Kong\\
	\textsuperscript{\rm 4}Shenzhen Institute of Advanced Technology
}

\usepackage{bibentry}

\begin{document}

\maketitle

\begin{abstract}
	Gaussian Splatting has emerged as a prominent 3D representation in novel view synthesis, but it still suffers from appearance variations, which are caused by
	various factors, such as modern camera ISPs, different time of day, weather conditions, and local light changes.
	These variations can lead to floaters and color distortions in the rendered images/videos. 
	Recent appearance modeling approaches in Gaussian Splatting are either tightly coupled with the rendering process, hindering real-time rendering, or they only account for mild global variations, performing poorly in scenes with local light changes. 
	In this paper, we propose DAVIGS, a method that decouples appearance variations in a plug-and-play and efficient manner. 
	By transforming the rendering results at the image level instead of the Gaussian level, our approach
	can model appearance variations with minimal optimization time and memory overhead. 
	Furthermore, our method gathers appearance-related information in 3D space to transform the rendered images, thus building 3D consistency across views implicitly.
	We validate our method on several appearance-variant scenes,
	and demonstrate that it achieves state-of-the-art rendering quality with minimal training time and memory usage, without compromising rendering speeds. 
	Additionally, it provides performance improvements for different Gaussian Splatting baselines in a plug-and-play manner.
	
\end{abstract}

\section{Introduction}
\begin{figure}[t]
	\centering
	\includegraphics[width=\columnwidth]{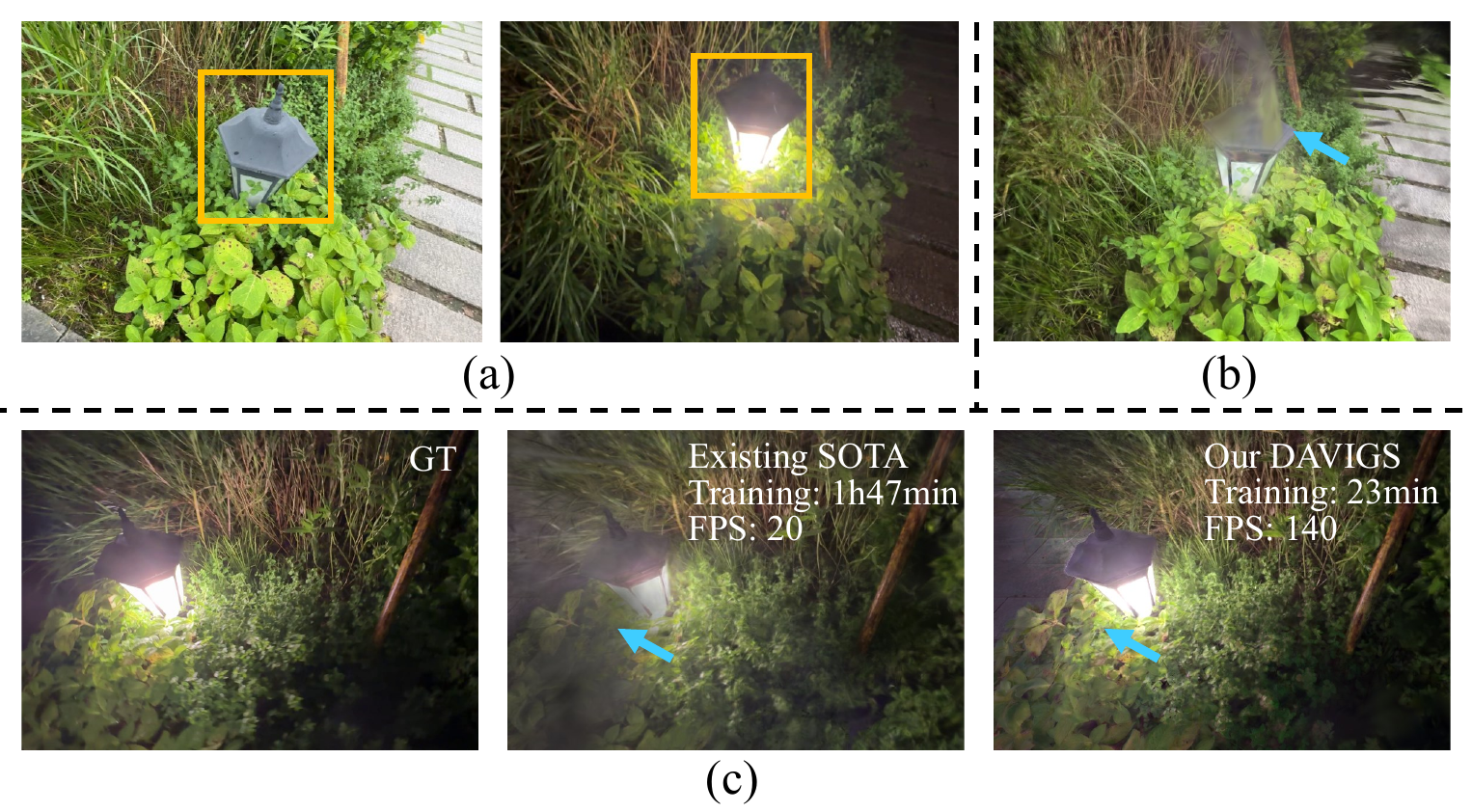} 
	\caption{(a) There are often global and local (orange box) appearance variations in real-world captures for 3D reconstruction. (b) Such appearance variations may lead to floaters. (c) Compared to existing SOTA methods, our DAVIGS achieves advanced reconstruction results with faster optimization and rendering.}
	\label{fig:teaser}
\end{figure}
As a promising representation in novel view synthesis, Gaussian Splatting (GS) enables high-quality reconstruction and real-time rendering.
Since the introduction of 3DGS \cite{kerbl20233d}, this representation has evolved into several variants, including Mip-Splatting \cite{Yu2024MipSplatting}, 2DGS \cite{huang20242d}, and GOF \cite{Yu2024GOF}, among others \cite{dai2024high,yang2024deformable,yu2024cogs,scaffoldgs,niemeyer2024radsplat,zhao2024badgaussians}.
These methods have been applied to various tasks, such as avatar creation \cite{li2024animatablegaussians, moreau2024human, kocabas2024hugs, abdal2024gaussian, qian2024gaussianavatars, zheng2024gpsgaussian,jiang2024hifi4g, liu2024humangaussian}, autonomous driving \cite{zhou2024drivinggaussian, yan2024street,zhao2024tclc,Zhou_2024_CVPR}, 3D asset generation \cite{chen2024text, tang2023dreamgaussian, yi2023gaussiandreamer,tang2024lgm,ling2024alignyourgaussians,zhou2024dreamscene360}, and SLAM \cite{keetha2024splatam, Matsuki:Murai:etal:CVPR2024,yan2024gs,hhuang2024photoslam}.
However, the impressive performance of GS relies on strict multi-view consistency, which is often compromised by appearance variations in real-world captures. Many factors, such as different camera image signal processings (ISPs), time of day, weather conditions, and local light changes can lead to noticeable alterations in the captured images, resulting in floaters and color distortions in the reconstructed scenes, as shown in Fig.~\ref{fig:teaser}.

\begin{figure}[t]
	\centering
	\includegraphics[width=\columnwidth]{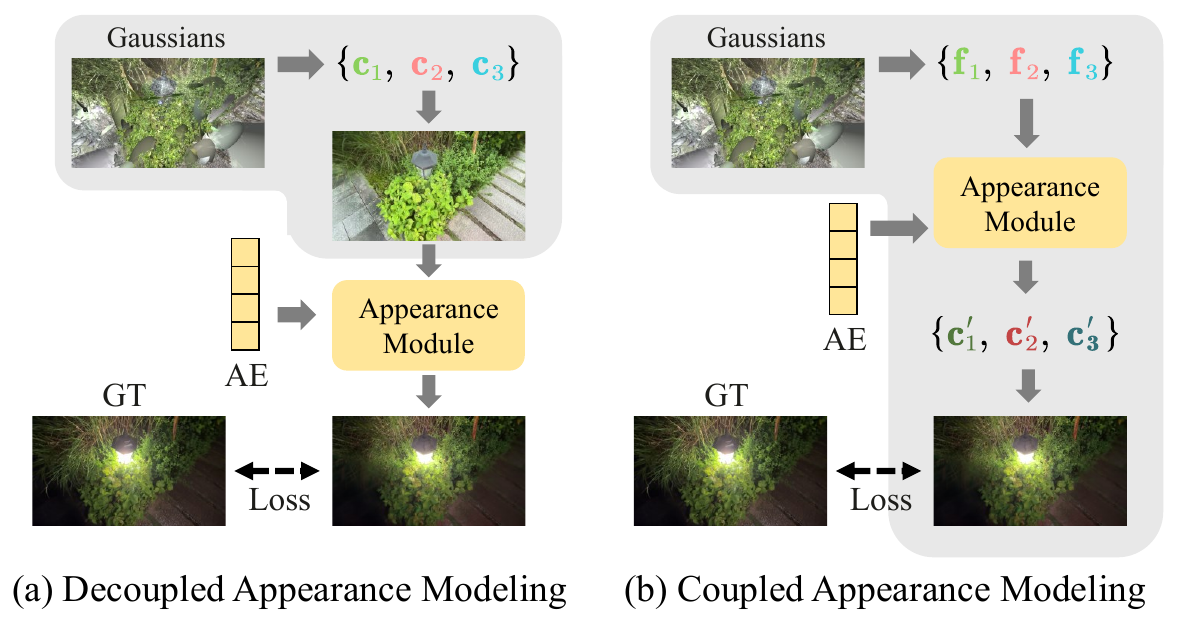} 
	\caption{\textbf{Different appearance modelings.} The gray parts represent the rendering process in Gaussian Splatting, where $\mathbf{f}$ denotes the features of Gaussian primitives, and $\mathbf{c}$ denotes the colors. (a) Decoupled appearance modeling applies transformations to rendering results at the image level, which can be discarded after optimization. (b) Coupled appearance modeling addresses appearance variations by manipulating the features of Gaussian primitives, making it coupled with the rendering process.}
	\label{fig:decouple}
\end{figure}

Previous appearance modeling methods \cite{martinbrualla2020nerfw, chen2022hallucinated, lin2024vastgaussian, zhang2024gaussian} typically optimize for an average scene appearance, followed by view-dependent transformations to match the actual appearances from different viewpoints. In NeRF-based methods \cite{martinbrualla2020nerfw, chen2022hallucinated}, this transformation occurs at the sampled points and is supervised in a pixel-wise manner. Appearance embeddings are attached to the point features and fed into a multi-layer perceptron (MLP) to generate view-dependent colors. 
GS-based methods \cite{lin2024vastgaussian, zhang2024gaussian} use a discrete 3D representation and are optimized in a frame-wise manner. Their appearance modeling can be categorized into two kinds: decoupled and coupled, as shown in Fig.~\ref{fig:decouple}.

The decoupled methods \cite{lin2024vastgaussian, darmon2024robust, niemeyer2024radsplat} perform transformations directly at the image level. These methods are advantageous due to their low computational and video memory requirements.
Additionally, the decoupling nature allows them to be discarded after optimization, thereby not impacting rendering speed. However, their primary drawback is the limited expressiveness, as existing decoupled methods are only suitable for mild and global changes, and struggle with more complex variations.

The coupled methods \cite{zhang2024gaussian, kulhanek2024wildgaussians} model appearance variations in the 3D space, and achieve stronger expressiveness by transforming Gaussian properties, such as color and opacity. This direct manipulation of 3D representations provides larger capacities to handle complex appearance variations. However, they are less efficient during rendering, as they cannot function without their appearance modeling.
Moreover, they require changes of the Gaussian properties to store additional information.

In this paper, we present DAVIGS (Decoupling Appearance Variations In GS), a novel appearance modeling method that integrates the strengths of both decoupled and coupled approaches. Our method transforms the rendering results at the image level, retaining the low-cost advantage of decoupled appearance modeling approaches without compromising rendering speed. To address the weakness of the decoupled methods, we incorporate 3D consistent features into the image transformation process, ensuring strong expressiveness.

We conceptualize appearance variation information as two components: global appearance embeddings associated with each image, and local features for different responses of various 3D locations, stored in multi-resolution hash grids.
These features are decoded by a lightweight MLP into an affine transformation matrix for each pixel in the image. 
To stabilize the optimization process and prevent unpleasant colors in the reconstructed scene, we introduce an identity regularization term that constrains the transformation to be close to the identity transformation.
Additionally, we propose a cell-based query approach to minimize the additional optimization time introduced by our appearance modeling.

Our DAVIGS method can be adapted to a wide range of complex appearance variations and ensures 3D consistency across views.
Our contributions are summarized as follows:
\begin{itemize}
	\item We propose DAVIGS, a novel decoupled appearance modeling approach that effectively mitigates appearance variation-induced floaters and color distortions with low optimization time and video memory overhead, without slowing down rendering speeds.
	\item We introduce a transformation regularizer and a cell-based query method to ensure stable and efficient optimization of our decoupled appearance module.
	\item Experiments conducted on both our new dataset and existing datasets demonstrate that DAVIGS achieves state-of-the-art performance in rendering quality, optimization time, video memory, and rendering speeds. Besides, it is plug-and-play for various GS methods.
\end{itemize}

\section{Related Work}
\begin{figure*}[t!]
	\centering
	\includegraphics[width=0.9\textwidth]{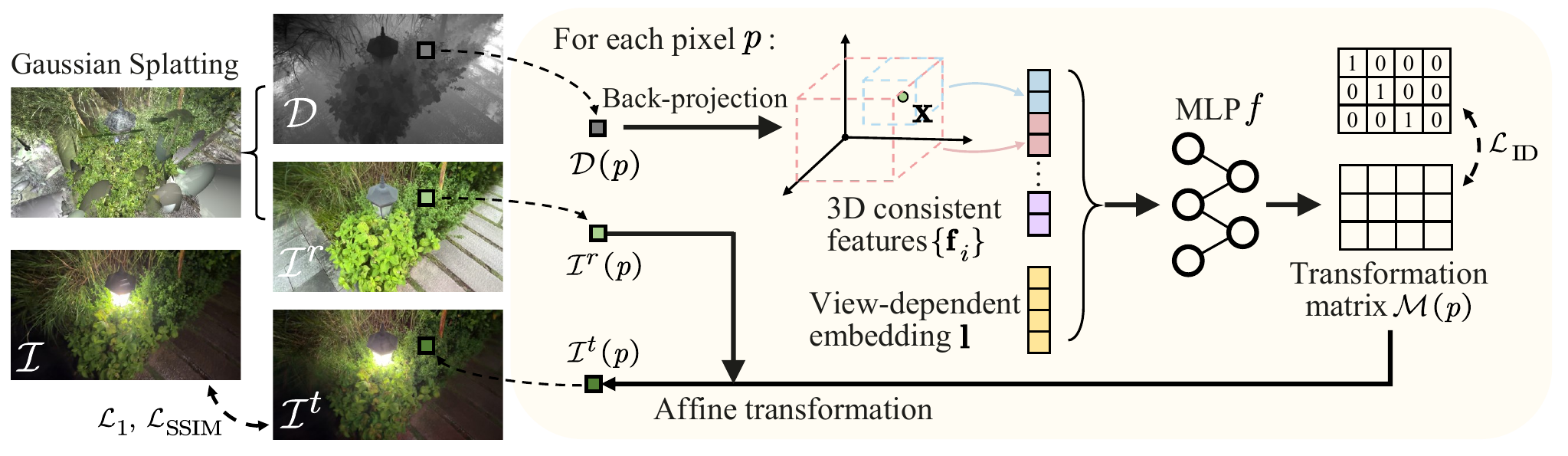} 
	\caption{\textbf{Overall pipeline of DAVIGS}. For each pixel $p$ in the rendered image $\mathcal{I}^r$, we calculate its 3D spatial position $\mathbf{x}$ by back-projecting with its depth $\mathcal{D}(p)$ in the depth map $\mathcal{D}$, and then look up the multi-resolution hash grids for its 3D consistent features $\{\mathbf{f}_i\}$. They are then concatenated with a view-dependent appearance embedding $\mathbf{l}$ and fed into an MLP $f$ to obtain a transformation matrix $\mathcal{M}(p) \in \mathbb{R}^{3\times 4}$, which is used to perform affine transformation on the color $\mathcal{I}^r(p)$ to obtain $\mathcal{I}^t(p)$. The losses $\mathcal{L}_1$ and $\mathcal{L}_\text{D-SSIM}$ are calculated between the transformed image $\mathcal{I}^t$ and the ground truth image $\mathcal{I}$. A regularization term $\mathcal{L}_\mathrm{ID}$ is applied to $\mathcal{M}(p)$ for constraining it close to the identity transformation matrix $\mathcal{M} _{\mathrm{ID}}$.}
	\label{pipeline}
\end{figure*}
There are two categories of appearance modeling approaches: decoupled and coupled, as shown in Fig.~\ref{fig:decouple}.
\subsection{Decoupled Appearance Modeling}
Decoupled appearance modeling applies transformations to rendering results at the image level.
VastGaussian \cite{lin2024vastgaussian} first proposes such an appearance modeling methodology.
It optimizes an appearance embedding (AE) for each image, which is concatenated to the downsampled rendered image and fed into a convolutional neural network (CNN) to decode a transformation map on the image. Robust Gaussian Splatting \cite{darmon2024robust} adopts a similar approach, but it directly optimizes a set of affine transformation parameters for each image. 
RadSplat \cite{niemeyer2024radsplat} introduces an exposure processing module that adjusts the rendered image based on the image's exposure duration and ISO setting.

The key advantage of these methods lies in their decoupling nature.
Once optimized, the appearance modules can be discarded without slowing down the rendering speed.
They are generally efficient during optimization, but their expressiveness is limited, as they can only adapt to relatively mild changes, such as camera auto-exposure and auto-white balance. 
Our proposed method, DAVIGS, belongs to this category but overcomes this limitation with 3D consistent features across views.

\subsection{Coupled Appearance Modeling}
Coupled appearance modeling operates directly on the 3D representations.
In NeRF-based methods \cite{martinbrualla2020nerfw, chen2022hallucinated}, AEs are attached to 3D sampled points in ray-marching and processed through an MLP to determine the points’ colors.
For GS-based methods, the manipulation is performed on Gaussian primitives. SWAG \cite{dahmani2024swag} optimizes hash grids and an MLP along with the attributes of 3D Gaussians. 
For each 3D Gaussian, the features of its position, its color and an AE are fed into the MLP to compute the color and an opacity residual.
Wild-GS \cite{xu2024wild} takes a more complex approach by feeding a global appearance embedding extracted from the image, local features obtained from the back-projection of the image, and intrinsic features of 3D Gaussians into an MLP to decode spherical harmonics (SH) coefficients.
GS-W \cite{zhang2024gaussian} uses adaptive sampling for each 3D Gaussian to sample appearance features from the feature maps extracted from the input image.
The features are blended and processed through an MLP along with a view direction to obtain the color.
WildGaussians \cite{kulhanek2024wildgaussians} extends the attributes of 3D Gaussians by adding an internal embedding, which is combined with the image’s AE and processed through an appearance MLP to obtain an affine mapping for determining view-dependent colors.

These coupled methods are capable of handling complex appearance variations, such as day-night transitions and different weather conditions.
However, they require additional attributes within Gaussian primitives and coupled MLP inference during rendering, which significantly increases optimization time and video memory, and slows down rendering speeds.

\section{Method}
\subsection{Preliminaries}
In this section, we present our novel DAVIGS method for efficient, plug-and-play appearance modeling in Gaussian Splatting. Since our primary experiments are conducted on 3D Gaussian Splatting (3DGS), which serves as the foundation for other GS variants, we begin with its basic preliminaries. 3DGS represents both geometry and appearance using a set of 3D Gaussians, each characterized by its position, anisotropic covariance, opacity, and spherical harmonic coefficients for view-dependent colors. During rendering, each 3D Gaussian is projected onto the image plane as a 2D Gaussian. These projected 2D Gaussians are then assigned to different tiles, sorted, and alpha-blended into the final image through a point-based volume rendering process.

The dataset used for scene optimization consists of a sparse point cloud and a set of training views. During the optimization process of 3DGS, one image $\mathcal{I}^r$ is rendered at each iteration using a differentiable rasterizer. This rendered image is then compared to the corresponding ground truth image $\mathcal{I}$ to compute the L1 loss and the D-SSIM loss. The overall loss function is given by:
\begin{equation}
	\mathcal{L}=(1-\lambda)\mathcal{L}_1(\mathcal{I}^r, \mathcal{I})+\lambda\mathcal{L}_\text{D-SSIM}(\mathcal{I}^r, \mathcal{I}),
	\label{eq:loss}
\end{equation}
where $\lambda$ is a hyperparameter. The gradients obtained through backpropagation of this loss are used to optimize the properties of the 3D Gaussians. This optimization process is interleaved with adaptive point densification, which is triggered when the cumulative gradient of a point reaches a predefined threshold.

\subsection{DAVIGS}
We introduce DAVIGS to solve the appearance variation problem in novel view synthesis, which includes global changes (e.g., image ISPs and daylight) and local changes (e.g., street lamps at night).
To achieve decoupling, we model appearance variations at the image level rather than at the Gaussian level. 
We use globally shared appearance embeddings (AEs) for each view, and learn distinct transformations for local changes of individual pixels.
To query 3D consistent local features, we compute the 3D position of each pixel,
and then calculate the transformation based on both the global AE and the local spatial features at that pixel.

\subsubsection{Architecture}
As illustrated in Fig.~\ref{pipeline}, we encode the global appearance of each view using an AE vector $\mathbf{l}\in \mathbb{R}^A$, store the local spatial features using multi-resolution hash grids $\{\mathbf{G}_i\}^{L-1}_{i=0}$ with $L$ resolution levels \cite{mueller2022instant}, and employ a lightweight MLP $f$ to learn affine transformation matrices. We transform the rendered image $\mathcal{I}^r$ at the image level. Specifically, during the differentiable rasterization process, we also render a depth map $\mathcal{D}$. For each pixel $\mathcal{I}^r(p)$ in the rendered image, we compute its corresponding 3D coordinate $\mathbf{x}\in \mathbb{R}^3$ via back-projection using its depth value $\mathcal{D}(p)$ and the camera parameters. Similar to grid-based methods in NeRF \cite{mueller2022instant}, we look up the surrounding voxels of coordinate $\mathbf{x}$ in the hash grids $\{\mathbf{G}_i\}^{L-1}_{i=0}$ and linearly interpolate the corner features to obtain the positional features of $\mathbf{x}$, denoted as $\{\mathbf{f}_i\in \mathbb{R}^{F}\}^{L-1}_{i=0}$. We then concatenate these positional features $\{\mathbf{f}_i\}^{L-1}_{i=0}$ with the corresponding appearance embedding $\mathbf{l}\in \mathbb{R}^A$ of the current view:
\begin{equation}
	\mathbf{f} = \mathbf{f}_0 \oplus \mathbf{f}_1 \oplus \cdots \oplus \mathbf{f}_{L-1} \oplus \mathbf{l}.
	\label{eq:concatenate}
\end{equation}
The resulting vector $\mathbf{f} \in \mathbb{R}^{LF+A}$ is then fed into the MLP $f$:
\begin{equation}
	f(\mathbf{f}) = \mathbf{v}.
	\label{eq:concatenate2}
\end{equation}
The output vector $\mathbf{v} \in \mathbb{R}^{12}$ is reshaped into an affine transformation matrix $\mathcal{M}(p)\in \mathbb{R}^{3\times 4}$. Finally, we apply this affine transformation to the RGB color of the pixel:
\begin{equation}
	\mathcal{I}^t(p) = \mathcal{I}^r(p) \mathcal{M}(p),
	\label{eq:affine}
\end{equation}
where $\mathcal{I}^t$ is the transformed image. This appearance modeling method encodes the global view-dependent appearance information in the appearance embeddings, while the local responses of different locations on the scene’s surface to the appearance features are sparsely stored in the hash grids. This approach ensures 3D consistency in both global and local appearances in the reconstructed scene. 
Since our approach transforms the rendered results at the image level, the module is plug-and-play for various GS methods and can be discarded after optimization without slowing down rendering speeds.

\subsubsection{Optimization}
Without regularization of the appearance learning process, the Gaussian primitives tend to adjust their colors instead of allowing the appearance model to optimize its parameters to fit the appearance variations. Although the final optimized scene may also converge to an average appearance, it might not be visually pleasing; e.g., a homogeneous surface could become blotchy. To address this, we introduce a regularization term $\mathcal{L}_\mathrm{ID}$ on the affine transformation matrix $\mathcal{M}(p)$ to ensure that it remains close to the identity transformation matrix $\mathcal{M} _{\mathrm{ID}}$:
\begin{equation}
	\mathcal{L}_\mathrm{ID} =\mathcal{L} _1\left( \mathcal{M} _{\mathrm{ID}}, \mathcal{M} (p) \right),\ \ \mathcal{M} _{\mathrm{ID}}=\left[ \begin{matrix}
		1&		0&		0&		0\\
		0&		1&		0&		0\\
		0&		0&		1&		0\\
	\end{matrix} \right] .
	\label{eq:id_loss}
\end{equation}
The final loss function is:
\begin{equation}
	\mathcal{L}=(1-\lambda_1)\mathcal{L}_1(\mathcal{I}^t, \mathcal{I})+\lambda_1\mathcal{L}_\mathrm{D-SSIM}(\mathcal{I}^t, \mathcal{I}) + \lambda_2 \mathcal{L}_\mathrm{ID},
	\label{eq:loss}
\end{equation}
where $\lambda_1$ and $\lambda_2$ are hyperparameters.
Although $\mathcal{I}^r$ is not directly supervised by the ground truth image $\mathcal{I}$, it is still properly constrained in an indirect manner, thanks to the regularization term $\mathcal{L}_\mathrm{ID}$ which penalizes the difference between $\mathcal{I}^t$ and $\mathcal{I}^r$.

\subsubsection{Cell-Based Query}
\begin{figure}[t]
	\centering
	\includegraphics[width=0.9\columnwidth]{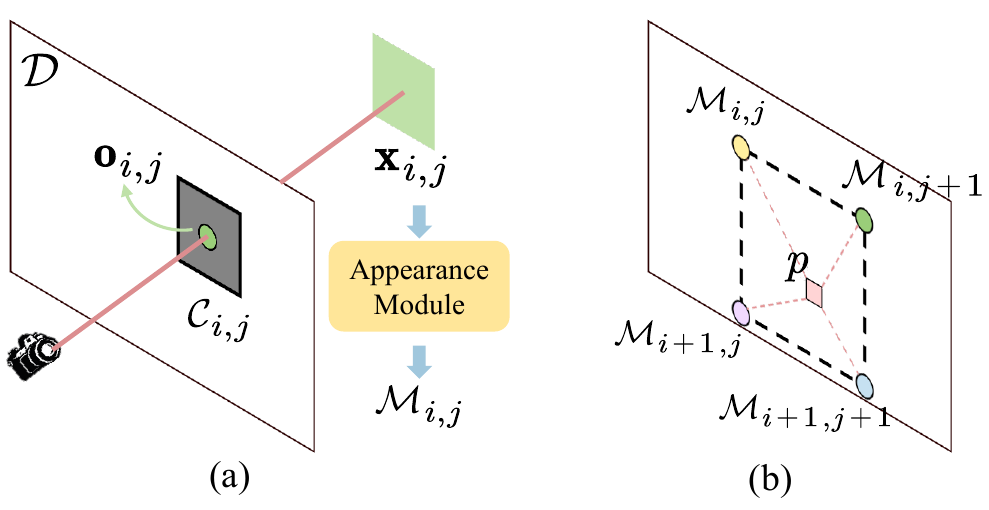} 
	\caption{\textbf{Cell-based query}. (a) We divide the depth map $\mathcal{D}$ into non-overlapping cells. For cell $\mathcal{C}_{i, j}$ with mean depth $\mathcal{D}_{i,j}$, we perform back-projection on its center $\mathbf{o}_{i,j}$ to obtain spatial coordinate $\mathbf{x}_{i,j}$. We use $\mathbf{x}_{i,j}$ to query the appearance module to obtain the transformation matrix $\mathcal{M}_{i,j}$. (b) For pixel $p$, we transform its color with the bilinearly interpolated affine transformation matrix.}
	\label{fig:tile}
\end{figure}

Querying the appearance model for every pixel is time-consuming, so we implement a cell-based query to compute the affine transformation matrix for one cell at a time. Specifically, as shown in Fig.~\ref{fig:tile}, let the resolution of the rendered image be $H\times W$; we divide it into non-overlapping cells in 2D with step size $s$, each of size $\frac{H}{s}\times \frac{H}{s}$. We then take the average of the depth map $\mathcal{D}$ within each cell indexed by $i, j$ (denoted as $\mathcal{D}_{i,j}$) and use the cell's central position $\mathbf{o}_{i,j}=\left( \frac{H}{s}i+\frac{H}{2s},\frac{W}{s}j+\frac{W}{2s} \right)$ to compute its 3D position via back-projection, denoted as $\mathbf{x}_{i,j}$. We use $\mathbf{x}_{i,j}$ to carry out the query in our appearance module to obtain the affine transformation matrix for the cell, denoted as $\mathcal{M}_{i,j}$. For any pixel $p$ in the rendered image, we use the affine transformation matrices of its neighboring cells to perform bilinear interpolation to obtain its affine transformation $\mathcal{M}(p)$. This cell-based query effectively reduces the optimization time. Given that spatially adjacent pixels tend to have similar appearance responses, this approach allows for faster optimization with minimal compromise in rendering quality.

\section{Experiments}
\subsection{Experimental Setups}

\subsubsection{Implementation Details}
In our main experiments, DAVIGS is implemented based on 3DGS. The MLP $f$ contains two hidden layers with widths of $128$ and $64$, respectively. The hyperparameter $\lambda_1$ is set to $0.2$, while $\lambda_2$ follows a cosine annealing function, warming up linearly from $0$ to $0.3$ in the first $5$k iterations and eventually decaying to $0.2$. For the depth used in back-projection, we employ the mean depth \cite{huang20242d} calculated during differentiable rasterization.
For cell-based query, we use a cell size of $8$.

\subsubsection{Datasets}
\begin{figure*}[t!]
	\centering
	\includegraphics[width=\textwidth]{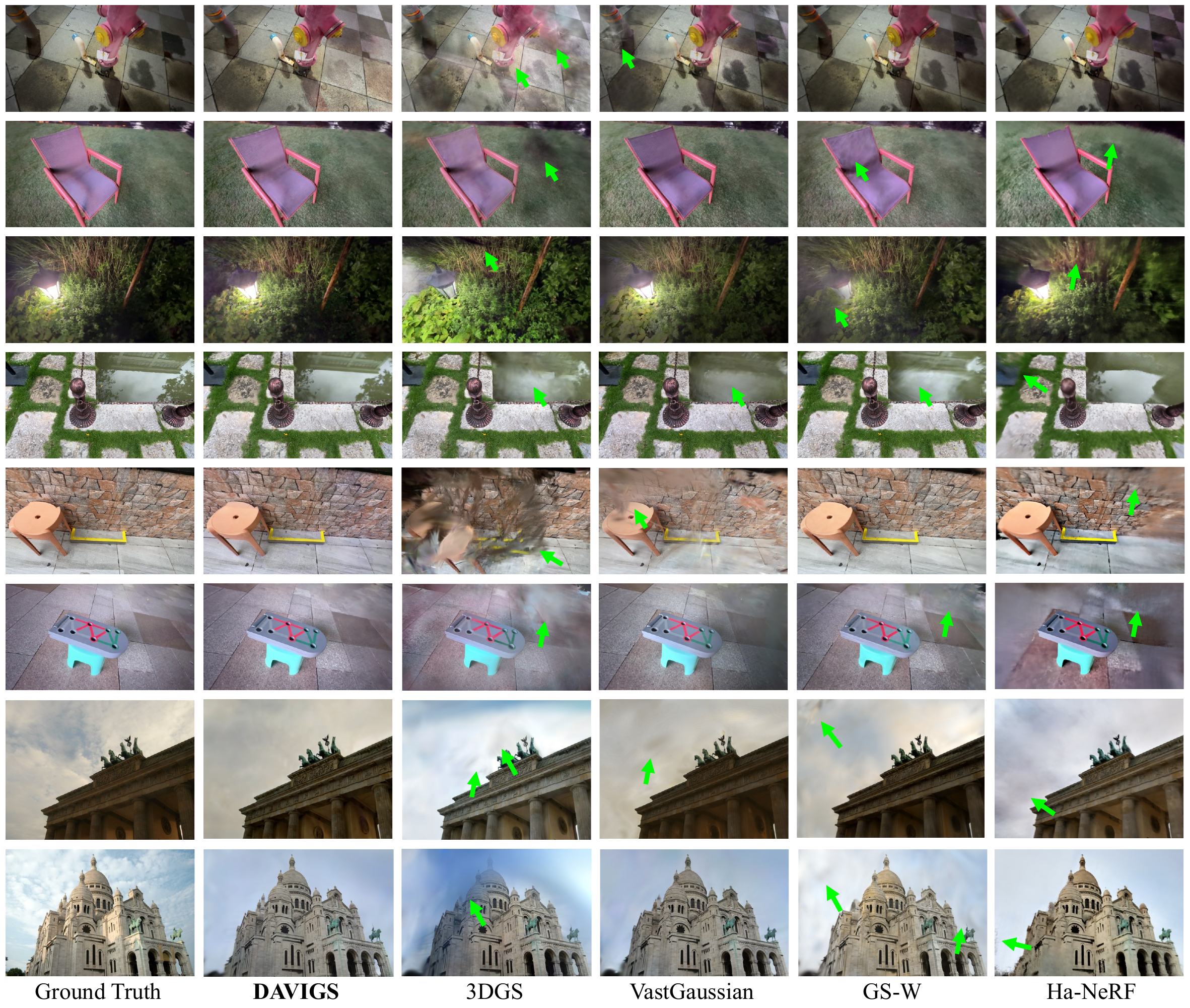} 
	\caption{Qualitative comparison between DAVIGS and previous work. Floaters and other artifacts are pointed out by arrows.}
	\label{fig:compare}
\end{figure*}

Our DAVIGS method focuses on decoupling appearance variations to alleviate floaters and color distortions. Existing datasets for evaluating this task are limited: some contain only global and mild appearance variations, such as the Mill-19 dataset \cite{Turki_2022_CVPR}, which primarily includes camera ISP variations; others contain other distractors, like the PhotoTourism dataset \cite{jin2021image}, which includes both appearance changes and transient objects, alongside with varying image resolutions and qualities. To facilitate a fair comparison, we create a high-quality \textit{GLAV dataset} comprising seven scenes that exhibit both \textbf{G}lobal and \textbf{L}ocal \textbf{A}ppearance \textbf{V}ariations. The images are captured with an iPhone 15 Pro camera at 1080p resolution and do not have any distractors other than appearance variations. The camera poses and sparse point clouds are estimated using COLMAP \cite{schoenberger2016sfm}. Additionally, we also evaluate our method on the PhotoTourism dataset, as it is widely used by other methods.

\subsubsection{Metrics} 
We evaluate rendering quality using PSNR, SSIM, and LPIPS, and assess efficiency using optimization time, VRAM usage, and rendering speed. Following the common practice, we use the left half of the test images during training to optimize the appearance-related parameters and reserve the right half for testing.
All the metrics are tested on a single Tesla V100 GPU.

\subsubsection{Compared Methods}
We compare DAVIGS with several other methods, including NeRF-W \cite{martinbrualla2020nerfw}, Ha-NeRF \cite{chen2022hallucinated}, VastGaussian \cite{lin2024vastgaussian}, Gaussian-W \cite{zhang2024gaussian}, and WildGaussians \cite{kulhanek2024wildgaussians}. For NeRF-based methods, we follow the default parameter settings of Ha-NeRF, training for $20$ epochs with a batch size of $1024$. For GS-based methods, we optimize the models for $30$k iterations, keeping other parameters the same as those in the original papers.

\subsection{Result Analysis}
As shown in Tab.~\ref{tab:davigs} and Fig.~\ref{fig:compare}, DAVIGS demonstrates superior rendering quality in comparison with other methods. 1)~Compared to \textit{vanilla 3DGS}, it is slightly slower during training, but has significant improvement in rendering quality. Since it reduces the floaters caused by appearance variances, it is slightly faster in rendering. Despite incorporating additional modules, DAVIGS requires less VRAM than 3DGS. 
2) Compared to the \textit{coupled appearance modeling methods} (GS-W, GS-W with cached features denoted as GS-W-c, and WildGS), DAVIGS has substantial advantages in terms of optimization time and VRAM usage since we transform the rendering results at the image level. The decoupled nature of DAVIGS also results in a much faster rendering process. 
3) Compared to another \textit{decoupled appearance modeling method} (VastGS), ours provides better 3D consistency, effectively resolving floaters and color distortions, leading to significant improvements in rendering quality. On the PhotoTourism dataset used by other methods, DAVIGS also achieves comparable quality with minimal training time and VRAM usage (see Tab.~\ref{tab:photo}).

\begin{table}[t!]
	\begin{center}
		\resizebox{\linewidth}{!}{
			\begin{tabular}{l|ccc|ccc}
				\toprule
				\multicolumn{1}{l|}{Dataset} & \multicolumn{6}{c}{GLAV} \\
				\midrule
				Metrics & PSNR$\uparrow$ & SSIM$\uparrow$ & LPIPS$\downarrow$ &  Training$\downarrow$ & VRAM$\downarrow$  & FPS$\uparrow$ \\
				\midrule
				NeRF-W   & 14.13 & 0.470 & 0.496 & 69h48min & 8.3G & $<$1   \\
				Ha-NeRF & 21.09 & 0.613 & 0.381 & 82h49min & 7.7G & $<$1 \\
				3DGS   & 22.80 & 0.753 & 0.221 & \tb{16min} & \underline{3.8G} & 134 \\
				GS-W  & \underline{27.23} & \underline{0.839} & \underline{0.141} & 1h47min & 8.1G & 24 \\
				GS-W-c  & 27.23 & 0.839 & 0.141 & 1h47min & 8.1G  & 84 \\
				WildGS  & 24.68 & 0.764 & 0.244 & 1h18min & 22.7G & 28 \\
				VastGS  & 23.21 & 0.782 & 0.186 & 28min & 4.7G & \underline{148} \\
				\midrule
				\tb{DAVIGS}  & \tb{28.31} & \tb{0.861} & \tb{0.134} & \underline{19min} & \tb{3.4G} & \tb{151} \\
				\bottomrule
			\end{tabular}		
		}
		\caption{Effectiveness (PSNR, SSIM, and LPIPS) and efficiency (training time, training VRAM usage, and rendering speed) evaluations on the GLAV dataset. 
			Our method achieves superior performance in all the metrics.}
		\label{tab:davigs}
	\end{center}
	\centering
\end{table}

\begin{table}[t!]
	\begin{center}
		\resizebox{\linewidth}{!}{
			\begin{tabular}{l|ccc|ccc}
				\toprule
				\multicolumn{1}{l|}{Dataset} & \multicolumn{6}{c}{PhotoTourism} \\
				\midrule
				Metrics & PSNR$\uparrow$ & SSIM$\uparrow$ & LPIPS$\downarrow$ &  Training$\downarrow$ & VRAM$\downarrow$  & FPS$\uparrow$ \\
				\midrule
				NeRF-W   & 20.78 & 0.799 & 0.208 & 114h20min & 8.3G & $<$1   \\
				Ha-NeRF & 21.41 & 0.793 & 0.178 & 271h13min & 7.7G & $<$1 \\
				3DGS   & 22.39 & 0.789 & 0.232 & \tb{20min} & \underline{6.7G} & \tb{130} \\
				GS-W   & \underline{23.81} & \tb{0.841} & \tb{0.135} & 2h28min & 13.9G & 18 \\
				GS-W-c  & 23.81 & 0.841 & 0.135 & 2h28min & 13.9G & 49 \\
				WildGS   & 22.81 & 0.824 & 0.158 & 1h40min & 18.8G & 22  \\
				VastGS   & 23.08 & 0.827 & 0.153 & 43min & 6.9G & 111 \\
				\midrule
				\tb{DAVIGS}  & \tb{23.95} & \underline{0.838} & \underline{0.148} & \underline{26min} & \tb{6.3G} &  \underline{123} \\
				\bottomrule
			\end{tabular}		
		}
		\caption{Despite some other methods apply additional modules (e.g., sky handling) for the PhotoTourism dataset, DAVIGS achieves comparable high quality with minimal training time and VRAM usage.}
		\label{tab:photo}
	\end{center}
	\centering
\end{table}

\subsection{Plug-and-Play on Different Baselines}
\begin{figure}[t]
	\centering
	\includegraphics[width=\columnwidth]{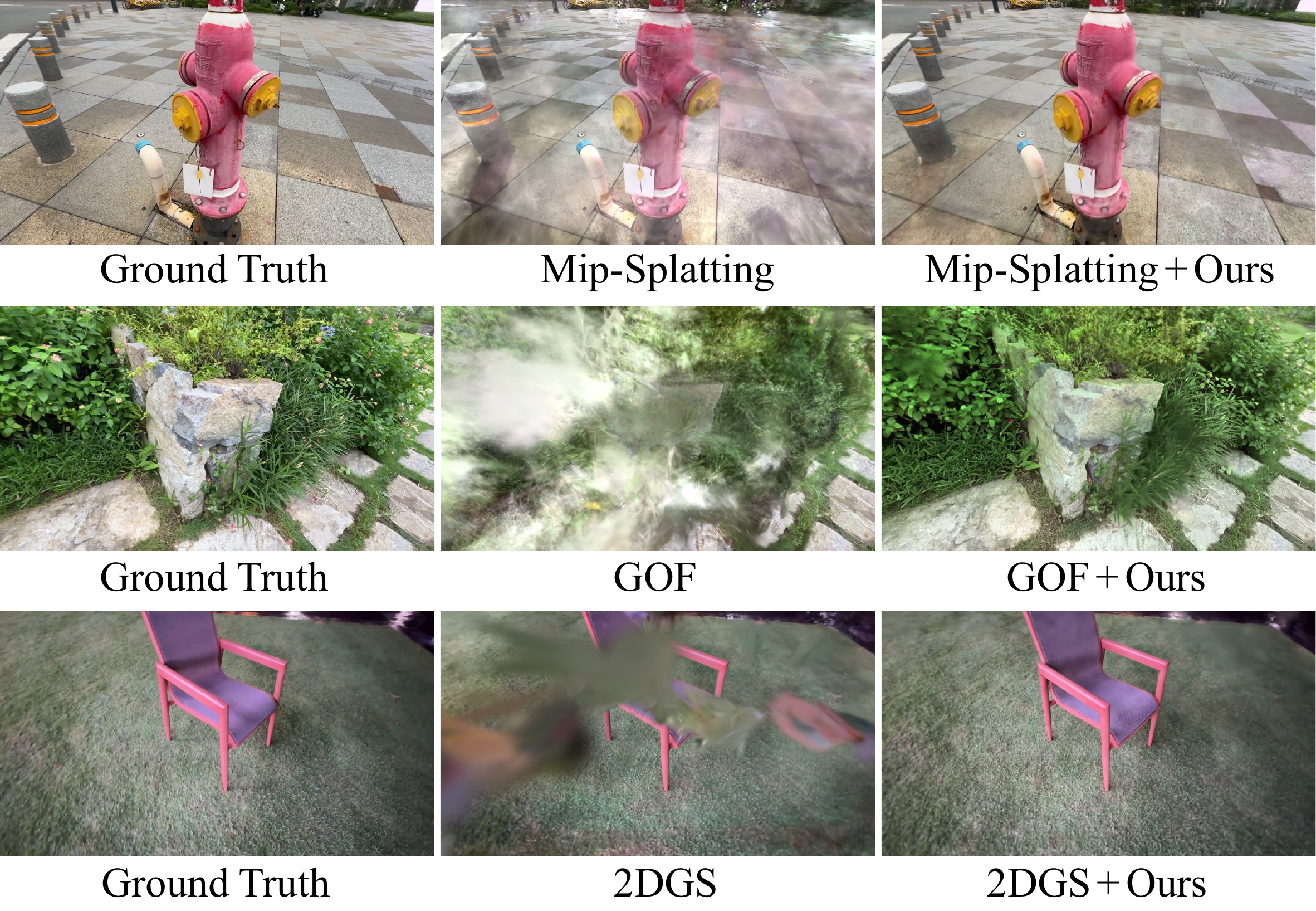} 
	\caption{DAVIGS can be plugged into different baselines to suppress floaters caused by appearance variations.}
	\label{fig:plug}
\end{figure}

To demonstrate the plug-and-play nature of our method, we implement DAVIGS on some other representative GS-based baselines, including Mip-Splatting \cite{Yu2024MipSplatting}, 2DGS \cite{huang20242d}, and GOF \cite{Yu2024GOF}. The parameters of our hash grids and MLP remain the same as those in the previous subsection, while the other parameters are kept as in the original papers. We compare these baselines with and without our appearance module on the GLAV dataset. As shown in Tab.~\ref{tab:plug} and Fig.~\ref{fig:plug}, integrating our module significantly improves the performance of these baselines. The integration requires minimal additional training time and even reduces VRAM usage due to the suppression of floaters, demonstrating the efficiency and plug-and-play capability of DAVIGS.

\begin{table}[t!]
	\begin{center}
		\resizebox{\linewidth}{!}{
			\begin{tabular}{l|ccccc}
				\toprule
				Metrics & PSNR$\uparrow$ & SSIM$\uparrow$ & LPIPS$\downarrow$ &  Training$\downarrow$ & VRAM$\downarrow$  \\
				\midrule
				3DGS   & 22.80 & 0.753 & 0.221 & \tb{16min8s} & 3.76G \\
				\tb{3DGS+Ours} & \tb{28.31} & \tb{0.861} & \tb{0.134} & 18min51s & \tb{3.39G} \\
				\midrule
				\midrule
				Mip-Splatting  & 22.46 & 0.748 & 0.295 & \tb{22min17s} & \tb{13.74G} \\
				\tb{Mip-Splatting+Ours} & \tb{28.31} & \tb{0.870} & \tb{0.130} & 24min52s & 14.17G \\
				\midrule
				\midrule
				GOF   & 22.38 & 0.737 & 0.231 & \tb{27min52s} & 18.11G \\
				\tb{GOF+Ours}   & \tb{28.21} & \tb{0.861} & \tb{0.139} & 28min9s & \tb{12.43G}  \\
				\midrule
				\midrule
				2DGS   & 22.70 & 0.746 & 0.229 & \tb{18min7s} & 6.16G \\
				\tb{2DGS+Ours} & \tb{27.76} & \tb{0.842} & \tb{0.158} & 20min & \tb{3.16G} \\
				\bottomrule
			\end{tabular}		
		}
		\caption{Our DAVIGS can be implemented into different baselines efficiently in a plug-and-play manner.}
		\label{tab:plug}
	\end{center}
	\centering
\end{table}

\subsection{Ablation Studies}
To assess the effectiveness of our introduced modules, we conduct ablation experiments on the GLAV dataset.
Following Mip-NeRF 360 \cite{barron2022mipnerf360} and VastGaussian \cite{lin2024vastgaussian}, we perform color correction before calculating the metrics to evaluate the quality of the average appearance (see the supplementary material).

\subsubsection{3D-Aware Information}
\begin{figure}[t]
	\centering
	\includegraphics[width=\columnwidth]{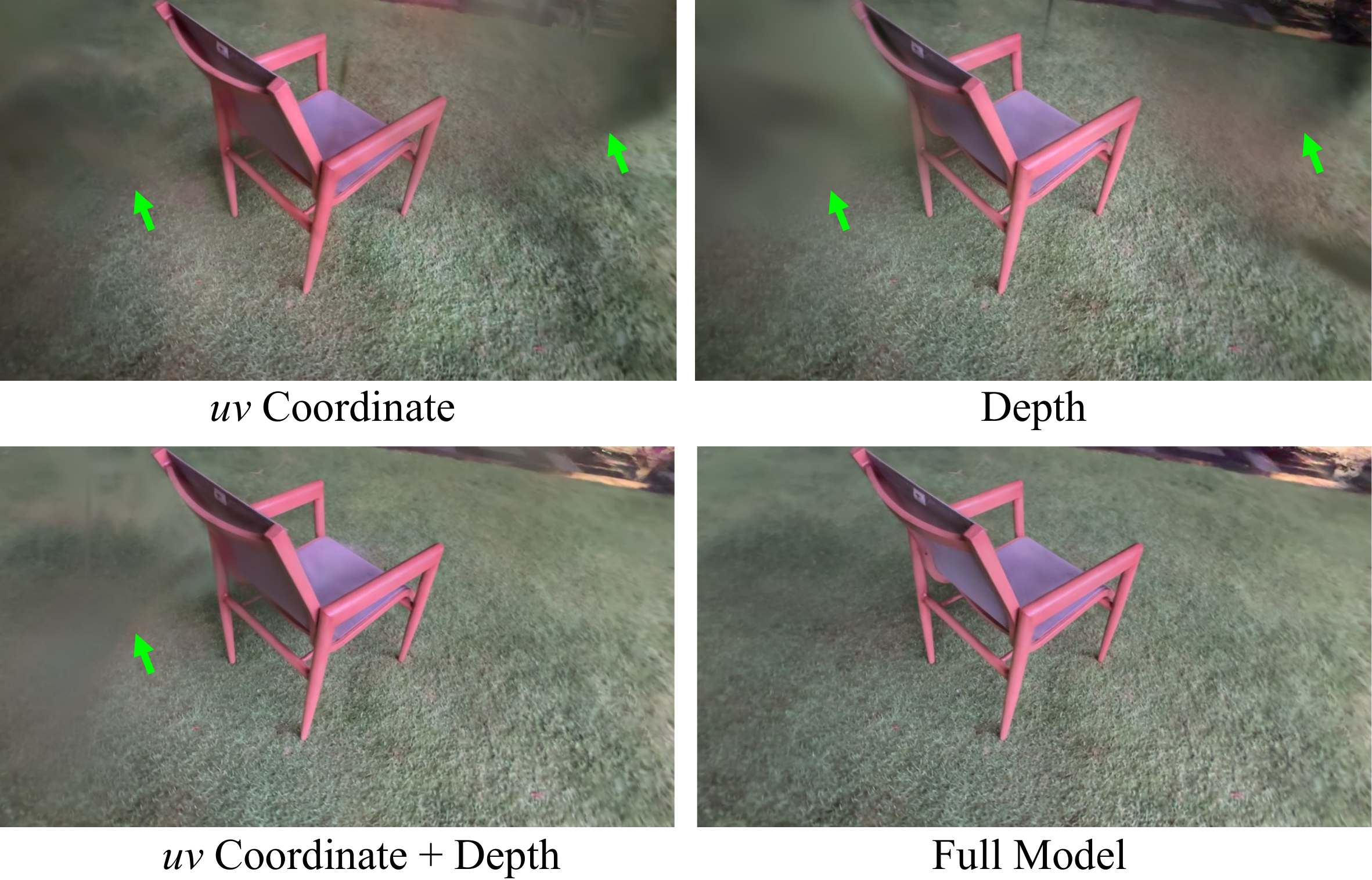} 
	\caption{The 3D features ($xyz$) in DAVIGS are necessary for suppressing floaters.}
	\label{fig:xyz}
\end{figure}

\begin{figure}[t]
	\centering
	\includegraphics[width=\columnwidth]{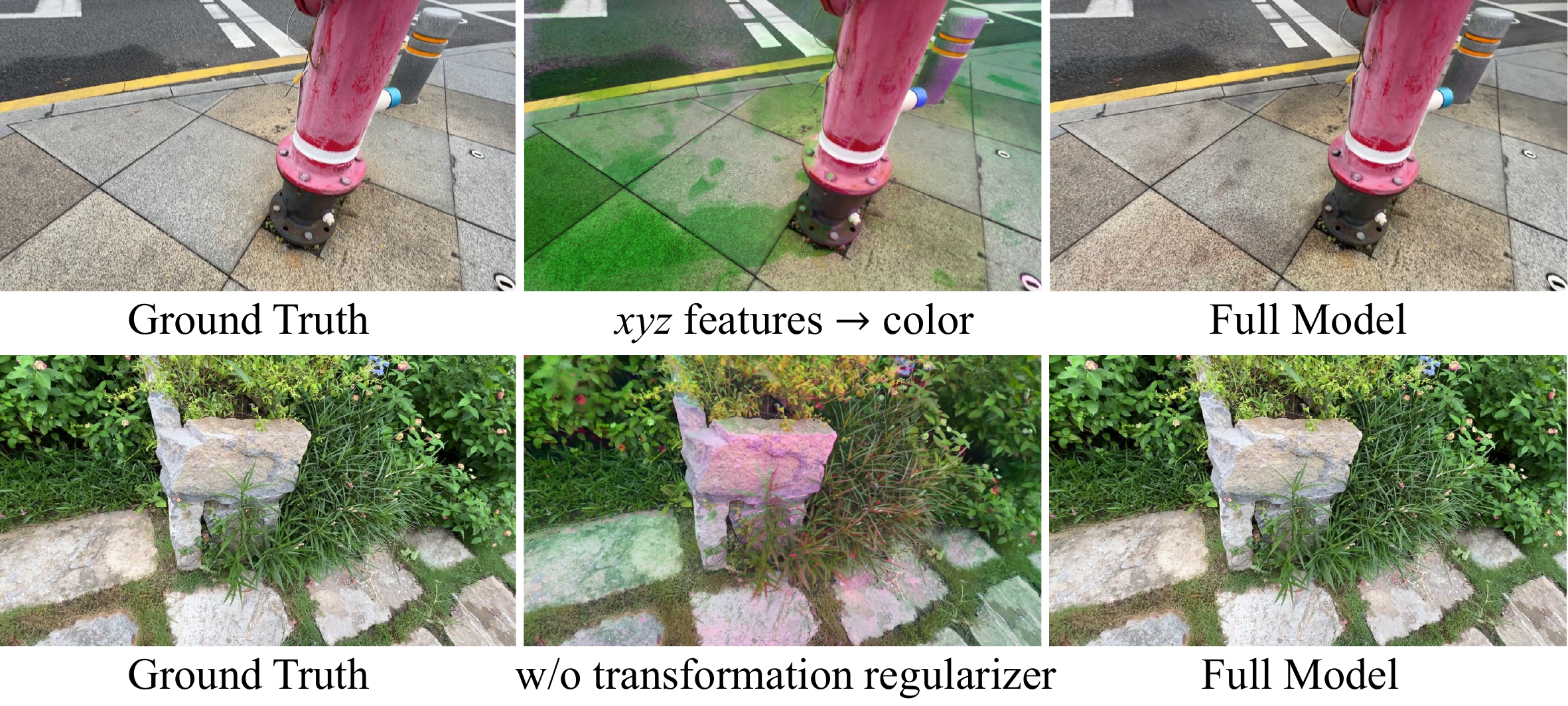} 
	\caption{Replacing the 3D features ($xyz$) with pixel colors or removing the transformation regularizer causes color distortions in the reconstructed scene.}
	\label{fig:color}
\end{figure}

To illustrate the importance of incorporating 3D-aware information in decoupled appearance modeling, we replace the 3D consistent features with four types of degenerated information: the $uv$-coordinates of each pixel in the rendered image, the depth, a combination of the previous two, and the RGB color. Each of these is embedded with positional encoding to match the length of our 3D features. As shown in Tab.~\ref{tab:3d} and Fig.~\ref{fig:xyz}, using pixel coordinates in the image or/and depth information significantly degrades the rendering quality compared to using the 3D information. When using only color information, 
the final optimized scene suffers from severe color distortions, as seen in Fig.~\ref{fig:color}.
\begin{table}[h!]
	\begin{center}
		\resizebox{\linewidth}{!}{
			\begin{tabular}{l|ccc}
				\toprule
				Model Setting & PSNR$\uparrow$ & SSIM$\uparrow$ & LPIPS$\downarrow$ \\
				\midrule
				$xyz$ $\rightarrow$ $uv$ coordinate & 21.76 & 0.798 & 0.252\\
				$xyz$ $\rightarrow$ depth & 20.04 & 0.788 & 0.268\\
				$xyz$ $\rightarrow$ $uv\ +\ $depth & 22.12 & 0.802 & 0.248\\
				$xyz$ $\rightarrow$ color & 21.58 & 0.765 & 0.317\\
				w/o $\mathcal{L}_\mathrm{ID}$ & 21.33 & 0.767 & 0.298\\
				\midrule
				\textbf{Full Model} & \textbf{22.75} & \textbf{0.803} & \textbf{0.243} \\
				\bottomrule
			\end{tabular}		
		}
		\caption{Ablation on the introduction of the 3D features ($xyz$) and transformation regularization term $\mathcal{L}_\mathrm{ID}$.}
		\label{tab:3d}
		\vspace{-3mm}
	\end{center}
	\centering
\end{table}
\subsubsection{Transformation Regularlization}
As shown in Tab.~\ref{tab:3d} and Fig.~\ref{fig:color}, removing the transformation regularizer 
allows the appearance module more freedom in learning appearances, which results in rendering quality degradation and color distortions in the optimized scene.

\subsubsection{Different Cell Sizes}
\begin{figure}[t]
	\centering
	\includegraphics[width=\columnwidth]{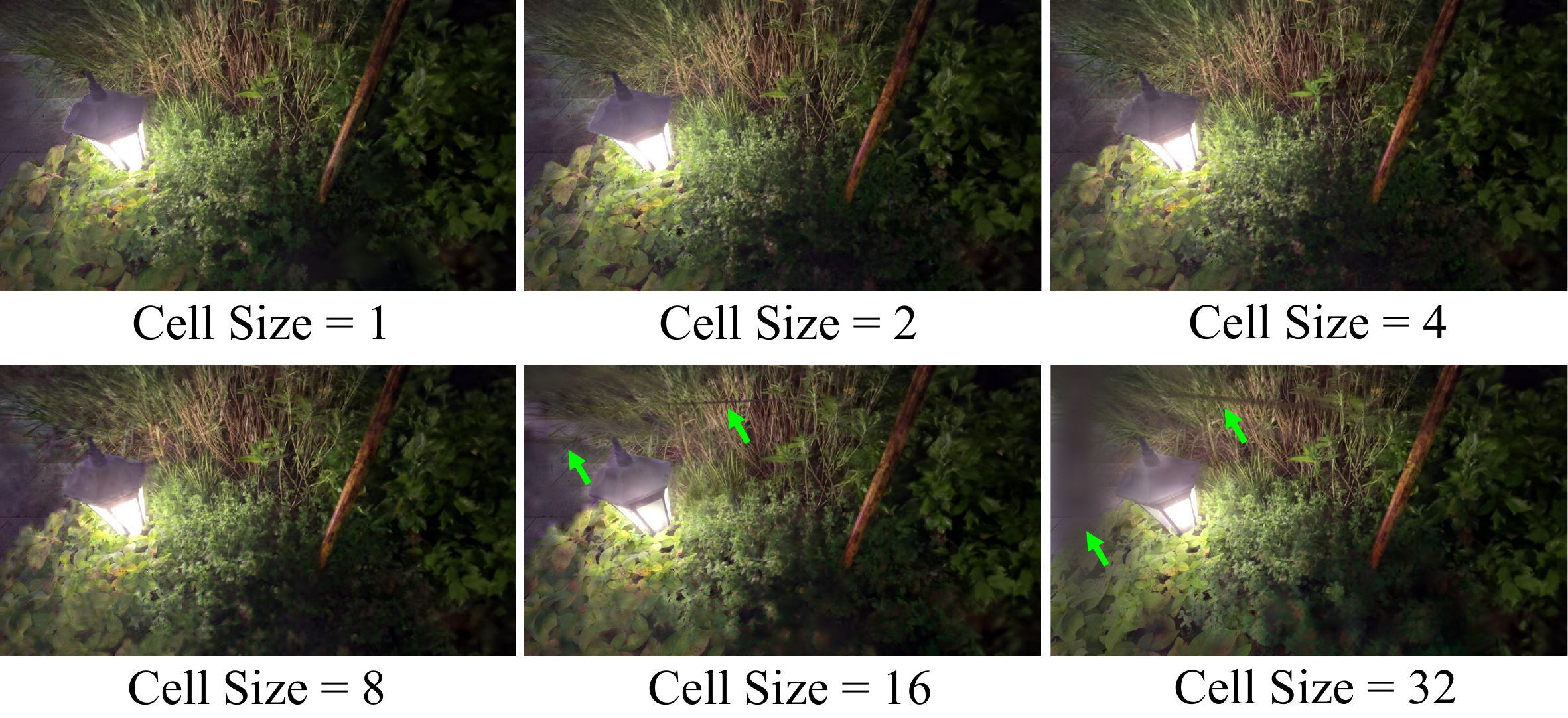} 
	\caption{Visual quality comparison of transformed images with different query cell sizes.}
	\label{fig:cell}
\end{figure}

We test the effect of different cell sizes on rendering quality and optimization speed (Tab.~\ref{tab:tile} and Fig.~\ref{fig:cell}). 
The visual quality of the transformed images decreases as the cell size increases, because a smaller cell size allows for more fine-grained transformations. 
When the cell size is greater than $8$, floaters start to appear in the scene.
For cell sizes of $8$ or smaller, the optimization time decreases as the cell size increases due to fewer queries in the appearance model. However, for cell sizes larger than $8$, the optimization time increases instead. This is because when the cell size increases, the gradients of the appearance model’s parameters are back-propagated to more Gaussian primitives, which causes an increase in time that outweighs the time saved by the cell-based query.

\begin{table}[t!]
	\begin{center}
		\resizebox{\linewidth}{!}{
			\begin{tabular}{c|cccc}
				\toprule
				Cell Size &  PSNR$\uparrow$ & SSIM$\uparrow$ & LPIPS$\downarrow$ &  Training$\downarrow$ \\
				\midrule
				1 & 28.24 & 0.858 & 0.134 & 28min35s \\
				2 & 28.35 & 0.861 & 0.134 & 20min36s  \\
				4 & 28.33 & 0.861 & 0.135 & 18min59s  \\
				8 & 28.31 & 0.861 & 0.134 & 18min51s  \\
				16 & 28.18 & 0.862 & 0.136 & 19min17s \\
				32 & 27.76 & 0.858 & 0.140 & 22min23s \\
				\bottomrule
			\end{tabular}		
		}
		\caption{Effect of different query cell sizes.}
		\label{tab:tile}
		\vspace{-3mm}
	\end{center}
	\centering
\end{table}

\section{Conclusions and Limitations}
In this paper, we introduce DAVIGS, an efficient and plug-and-play decoupled appearance modeling method that effectively alleviates floaters and color distortions caused by image appearance variations in novel view synthesis. Our approach achieves state-of-the-art rendering quality with minimal optimization time and video memory overhead. Besides, since the introduced module can be discarded after optimization, it does not slow down rendering speeds. 

However, when scaling up to larger scenes, our method may require larger multi-resolution hash grids and deeper MLPs. This can result in increased video memory usage, storage demand, and longer optimization time.
This scaling problem can be solved by some divide-and-conquer strategies \cite{lin2024vastgaussian, liu2024citygaussian}, and we leave it for future work.

\section{Acknowledgments}
This work was partly supported by the National Natural Science Foundation of China (No. 62171251) and the Special Foundations for the Development of Strategic Emerging Industries of Shenzhen (No. KJZD20231023094700001).

\bibliography{aaai25}

\end{document}